\documentclass[]{spie}

\usepackage{amsmath,amssymb}
\usepackage{graphicx}
\usepackage{booktabs}
\usepackage{multirow}
\usepackage{microtype}
\usepackage{hyperref}
\usepackage{cite}

\title{BINO: Encoder Centric Self Supervised Stereo With Native Pair Input}

\author[a]{Haokun Zhou}
\affil[a]{Imperial College London}
\authorinfo{Correspondence: Haokun Zhou, E-mail: haokun.zhou25@imperial.ac.uk}

\begin{document}
\maketitle

\newcommand{\ours}{\textsc{Ours}}
\newcommand{\fullours}{\texttt{OURS}}


\begin{abstract}
Stereo needs features that preserve fine cross view correspondence rather than only semantic similarity. Recent self supervised vision models transfer well, but they are not built for this goal, and geometry focused methods often rely on a binocular decoder or another explicit linkage module during pretraining. BINO asks whether strong binocular structure can instead be learned inside a compact encoder. It does this by fusing the rectified pair at the input stage, forming stereo micro cell tokens, and using a row aware patch phase positional encoding. Training uses one view masked token only distillation together with occlusion and view specific appearance mismatch. In a strict low resource setting with pretraining only on KITTI object, BINO gives the best frozen descriptor results under a no linkage probe among all compared baselines on proxy dense stereo, hard negative retrieval, and KITTI Stereo~2012 disparity. With the same lightweight stereo head for every encoder, it stays near CroCo~v2 while using a much smaller encoder. Supplementary transfer experiments on KITTI Stereo~2015 show the same qualitative trend. These results suggest that much of the cross view reasoning often assigned to a separate linkage module can be learned inside a compact and reusable encoder.
\end{abstract}

\section{Introduction}

Dense geometric tasks require feature representations that are not only robust to appearance change but also sensitive to view dependent spatial structure. This requirement is especially sharp in stereo, where the encoder must preserve correspondence cues under occlusion, photometric mismatch, and limited baseline. In recent years, self supervised vision models have become extremely strong as generic feature extractors. Methods based on masked image modeling and self distillation, such as MAE, iBOT, DINOv2, and DINOv3, produce robust features for recognition, dense prediction, and transfer \cite{He2022MAE,Zhou2021iBOT,Oquab2023DINOv2,Simeoni2025DINOv3}. However, these methods are not explicitly designed to learn dual view geometry. Their objectives are driven mainly by semantic reconstruction or global invariance rather than by the need to resolve spatial ambiguity between two synchronized viewpoints.

Cross view completion provides a more direct route toward geometric representation learning. CroCo formulates self supervision as predicting masked content in one view from visible content and a second viewpoint of the same scene \cite{Weinzaepfel2022CroCo}. CroCo~v2 strengthens this idea and shows that cross view completion can be scaled to stereo matching and optical flow by using a monocular encoder and a binocular decoder \cite{Weinzaepfel2023CroCov2}. These results establish a strong category of geometry oriented self supervision. At the same time, they leave open a central question. Is a dedicated binocular linkage module actually necessary during representation learning, or can sufficiently strong dual view spatial understanding emerge from an encoder alone?

This question matters for both scientific and practical reasons. Scientifically, linkage modules conflate two sources of performance, namely the quality of the representation itself and the power of the matcher layered on top of it. Practically, linkage modules increase training complexity, inference cost, and parameter count. A stereo oriented encoder that already captures useful binocular structure could be used in frozen form, paired with a lightweight downstream matcher, or reused in settings where a heavy binocular head is undesirable.

In this paper we study a binocular encoder whose central design principle is to push cross view structure into the encoder itself. The model fuses left and right images at the input level through pixel interleaving, uses stereo micro cell tokens, and is trained with cross view masked token distillation. The encoder does not rely on a binocular decoder, a cost volume, or another separate learned linkage module during pretraining. The same encoder can be queried in a deterministic duplicate view mode for descriptor extraction without introducing test time augmentation or an alternative inference path. The core question of the paper is therefore whether such an encoder can simultaneously satisfy three criteria: strong dual view spatial understanding, no reliance on a linkage network, and meaningful single image compatibility.

We answer this question in two stages. First, we construct a controlled benchmark that isolates dual view spatial understanding through encoder only correspondence metrics. This benchmark lets us ablate the architecture, the objective, and the training data distribution. Second, we evaluate on real stereo data in a strict low resource setting, where all encoders are pretrained from scratch on KITTI object and compared under a frozen no linkage protocol as well as under a shared lightweight stereo head. In the no linkage regime, our method is the strongest among all compared baselines on proxy dense stereo, hard negative retrieval, and real KITTI Stereo~2012 disparity. Under the shared head protocol, our model remains one of the two strongest encoders while using far fewer encoder parameters than CroCo~v2. Supplementary KITTI Stereo~2015 transfer experiments show the same qualitative trend in both the frozen no linkage and shared head settings, which suggests that the learned binocular latent is not specific to a single benchmark. These results support a clear conclusion: strong dual view spatial encoders can be learned without a heavy linkage network, and the quality of the encoder itself remains visible even when a shared downstream matcher is introduced.

\section{Related Work}

\paragraph{Single-view self-supervised visual pretraining.}
Recent large-scale self-supervised vision models have established strong generic visual features without labels. MAE showed that masked image reconstruction with an asymmetric encoder--decoder is highly effective for scalable representation learning \cite{He2022MAE}. iBOT introduced masked token prediction through self-distillation and highlighted the importance of patch-level supervision \cite{Zhou2021iBOT}. DINOv2 demonstrated that carefully scaled self-distillation can produce robust general-purpose visual features without supervision \cite{Oquab2023DINOv2}, while DINOv3 further pushed this line toward stronger dense features and larger-scale foundation models \cite{Simeoni2025DINOv3}. These methods are highly relevant baselines because they define the modern standard for self-supervised visual pretraining. At the same time, they are fundamentally single-view objectives. They do not directly force the encoder to resolve ambiguity between synchronized viewpoints, which is the central challenge in stereo geometry.

\paragraph{Self-supervised stereo and binocular depth learning.}
A substantial line of work learns stereo or depth directly from unlabeled binocular or multiview data through photometric consistency, epipolar structure, and mutual supervision. Flow2Stereo jointly learns optical flow and stereo by exploiting the geometric relation between the two tasks in stereoscopic videos \cite{Liu2020Flow2Stereo}. H-Net incorporates mutual epipolar attention inside a Siamese autoencoder for unsupervised stereo depth estimation \cite{Huang2022HNet}. Stereo Matching by Self-supervision of Multiscopic Vision further extends self-supervision beyond binocular pairs by using aligned multi-camera capture together with cross-photometric and uncertainty-aware mutual-supervision losses \cite{Yuan2021Multiscopic}. These methods are important precedents because they show that binocular self-supervision can produce strong geometric signals without dense labels. However, their main target is still end-to-end disparity or depth prediction. Our goal is different: rather than optimizing a dedicated stereo estimator, we ask whether a compact frozen encoder can acquire a reusable binocular spatial latent that remains strong across multiple readout protocols.

\paragraph{Monocular--binocular transfer and unified self-supervised depth models.}
Another line of work studies the reciprocal relation between monocular and stereo self-supervision. Reversing the Cycle distills monocular completion cues into stereo training in order to soften typical stereo artifacts \cite{Aleotti2020ReversingCycle}. Chen \emph{et al.}\ explicitly analyze the reciprocal relations between self-supervised stereo and monocular depth estimation \cite{Chen2021ReciprocalStereoMono}. More recently, TiO-Depth proposes a single model intended to serve both monocular and binocular self-supervised depth estimation \cite{Zhou2023TiODepth}. These works are especially relevant to our single-image compatibility claim, because they frame binocular and monocular supervision as mutually informative rather than disjoint. Our setting differs in emphasis. We do not aim to build a unified depth estimator with dedicated monocular and binocular prediction pathways. Instead, we preserve duplicated-view single-image compatibility within the same binocular encoder, while treating the core problem primarily as one of representation learning rather than end-to-end depth regression.

\paragraph{Cross-view completion, pairwise geometry pretraining, and explicit 3D prediction.}
CroCo introduced cross-view completion as a self-supervised task for 3D vision, where masked target content is reconstructed from another view of the same scene \cite{Weinzaepfel2022CroCo}. CroCo~v2 extended this idea to stereo and optical flow, showing that cross-view completion can scale effectively when paired with a monocular encoder and a binocular decoder \cite{Weinzaepfel2023CroCov2}. More recent pairwise geometry models such as DUSt3R and MASt3R go further by directly predicting pointmaps or matching-oriented dense features in 3D \cite{Wang2024DUSt3R,Leroy2024MASt3R}. These methods are highly relevant because they show that synchronized views can be used to learn powerful geometric priors. Our work differs in emphasis. We ask whether much of this binocular structure can be pushed into the encoder itself, without a binocular decoder, explicit pointmap regression, or other model-specific linkage module during pretraining.

\paragraph{3D-aware distillation and alternative supervision sources.}
A separate family of methods injects geometry through distillation or alternative supervision sources rather than through direct binocular latent prediction alone. Neural Feature Fusion Fields (N3F) distills self-supervised 2D image features into a scene-specific 3D neural field using differentiable rendering \cite{Tschernezki2022N3F}. KD-MVS introduces teacher--student distillation within self-supervised multi-view stereo \cite{Ding2022KDMVS}. 3D Distillation uses multi-view 3D information to improve self-supervised monocular depth estimation in difficult reflective regions \cite{Shi2023ThreeDDistillation}. NeRF-Supervised Deep Stereo uses neural rendering to synthesize stereo supervision and proxy depth from monocular capture \cite{Tosi2023NeRFSupervisedStereo}. These works demonstrate that 3D structure and distillation can substantially improve geometric learning, but they rely on additional 3D teachers, scene-specific radiance fields, rendered proxy labels, or explicit teacher--student reconstruction pipelines. Our method remains lighter-weight: the teacher and student share the same binocular encoder family, the supervision is slot-wise token distillation on raw stereo pairs, and no external 3D teacher or rendered labels are required.

\paragraph{Stereo matching and learned linkage.}
Classical stereo methods such as semi-global matching separate descriptor cost construction from spatial regularization \cite{Hirschmuller2008SGM}. Modern learned stereo systems typically go further by building explicit cross-view linkage through cost volumes, correlation layers, recurrent refinement, or binocular decoders. Recent foundation-style stereo systems continue this trend and achieve strong zero-shot performance by pairing rich encoders with powerful matching pipelines \cite{Wen2025FoundationStereo}. These models are important reference points, but they answer a different question than the one addressed here. Our goal is not to design the strongest end-to-end stereo matcher. Our goal is to evaluate the quality of the encoder itself. For that reason, our primary evaluation forbids learned linkage at inference time and uses only descriptor matching on frozen encoder features, while a shared-head benchmark is used separately to factor out downstream matcher design.

Unlike prior self-supervised stereo objectives that supervise end-to-end flow, disparity, photometric reconstruction, or 3D prediction \cite{Liu2020Flow2Stereo,Huang2022HNet,Yuan2021Multiscopic,Ding2022KDMVS,Wang2024DUSt3R}, our target is a slot-aligned token distribution in latent space.

\section{Method}

\subsection{Overview and design principle}

Given a rectified stereo pair $(L,R)\in[0,1]^{3\times H\times W}$, our goal is to learn a self supervised encoder that produces a strong \emph{binocular spatial latent} without relying on a binocular decoder, a cost volume, or another learned linkage module during pretraining. The central question is whether dual view spatial understanding can be pushed into the encoder itself, so that the resulting representation is useful both as a frozen descriptor field and as a compact latent for downstream stereo readout.

The proposed design is built around four choices. First, the two views are fused at the input level rather than kept as separate streams. Second, tokenization is performed on local \emph{stereo micro cells} that already contain information from both views. Third, positional structure is matched to rectified stereo through a row aware patch phase parameterization rather than a generic absolute 2D code. Fourth, training uses one view masked token only distillation, so the student must infer missing content from the complementary view. In addition, pretraining includes stereo aware nuisance perturbations such as occlusion and view specific appearance change, because these are precisely the conditions under which real correspondence becomes difficult.

The same pretrained encoder is later evaluated through three readout regimes. It can be exported as a frozen descriptor map without any learned matcher, paired with the same lightweight shared stereo head used for all baselines, or evaluated directly under its native pair input formulation with a common fused grid decoder. This organization mirrors the experimental structure of the paper.

\subsection{Early binocular fusion and stereo micro cell tokenization}

We begin by fusing the stereo pair into a single image like tensor
\[
X=\Phi(L,R)\in\mathbb{R}^{3\times H\times 2W}
\]
through horizontal pixel interleaving,
\[
X(:,:,2u)=L(:,:,u),\qquad X(:,:,2u+1)=R(:,:,u),
\]
for image column $u\in\{0,\dots,W-1\}$. A patch embedding of size $p_h\times p_w$ is then applied to $X$, with $p_w$ chosen to be even. Because the fused image alternates left and right pixels along the horizontal axis, each token covers a local fragment from the left view and a local fragment from the right view simultaneously. We refer to this fused receptive field as a \emph{stereo micro cell}.

This input design creates \emph{early binocular coupling}. In a two stream encoder, left and right tokens remain monocular until a later interaction stage. Here, every token is binocular from the beginning, so cross view structure is injected before the first transformer block. The encoder therefore operates over a sequence whose elementary units already contain local correspondence information.

The same architecture can also accept a degenerate duplicated input $(I,I)$ formed from a single image $I$. This preserves single image compatibility without introducing a separate inference branch, and later allows deterministic descriptor export from the same pretrained model.

\subsection{Row aware patch phase positional parameterization}

Rectified stereo has a highly structured geometry. Valid correspondences lie on the same epipolar row, while ambiguity is concentrated along the horizontal disparity direction. We therefore use a positional parameterization that reflects this structure.

Let the fused token grid be indexed by row $r$ and fused horizontal index $c$. We rewrite the fused horizontal coordinate as
\[
c=2p+q,
\]
where $p$ is the original patch column index and $q\in\{0,1\}$ is the local phase within the stereo micro cell. The pair $(p,q)$ expresses the fact that two adjacent fused tokens belong to the same original patch column but correspond to different within cell phases.

We define a learnable row embedding $a_r\in\mathbb{R}^d$ and initialize the token features as
\[
t^0_{r,c}=e_{r,c}+a_r,
\]
where $e_{r,c}$ is the patch embedding output. No separate absolute horizontal embedding is added outside attention. Instead, horizontal geometry is introduced by rotary position encoding applied inside attention using the 2D index pair $(r,p)$ rather than the raw fused index $(r,c)$.

This design is \emph{row aware} because row identity is explicit, and it is \emph{patch phase aligned} because the two phase tokens of the same stereo micro cell share the same patch column coordinate. In effect, the model encodes relative horizontal structure at the patch column level while preserving the local two phase organization of each binocular cell. This reduces the tendency to memorize absolute fused column identity, which is not the right inductive bias for disparity reasoning.

After the final transformer block, we form a de positioned representation
\[
\tilde t_{r,c}=t_{r,c}-a_r.
\]
This de positioned token state is used both before the projection head and when exporting descriptors. The purpose is to preserve the useful epipolar row prior inside the encoder while reducing the extent to which the final supervision or downstream readout can rely on trivial absolute row identity alone.

\subsection{One view masked token only distillation}

Pretraining uses an EMA teacher student framework. The teacher receives the full binocular input
\[
X^t=\Phi(L,R),
\]
while the student receives a version in which exactly one view is patch masked,
\[
X^s=\Phi(M(L),R)\quad\text{or}\quad X^s=\Phi(L,M(R)),
\]
where $M(\cdot)$ denotes random patch masking. The masked view is chosen at random. This asymmetry is essential: one view remains intact, so successful prediction requires the student to recover missing information from the complementary viewpoint rather than from trivial monocular continuity alone.

Let $\tilde t^t_j$ and $\tilde t^s_j$ denote teacher and student token representations at slot $j$, and let $h(\cdot)$ be the token projection head. The token logits are
\[
z^t_j=h(\tilde t^t_j),\qquad z^s_j=h(\tilde t^s_j).
\]
Following DINO style self distillation, the teacher distribution is centered and sharpened,
\[
p^t_j=\mathrm{softmax}\!\left(\frac{z^t_j-\mu}{\tau_t}\right),\qquad
p^s_j=\mathrm{softmax}\!\left(\frac{z^s_j}{\tau_s}\right),
\]
where $\mu$ is the running teacher center and $\tau_t,\tau_s$ are the teacher and student temperatures. The final pretraining loss is the uniform token level cross entropy
\[
\mathcal{L}_{\mathrm{tok}}
=
-\frac{1}{N}\sum_j p^t_j\cdot \log p^s_j.
\]

The final model uses \emph{token only} supervision rather than a mixed token+CLS objective. This choice is deliberate. Our target is dense dual view spatial understanding, not primarily global invariance, and the controlled ablations later show that adding a global branch weakens dense matching quality.

This objective is best interpreted as \emph{slot wise conditional inference in latent space}. The student is trained to reproduce the teacher's latent prediction at the same spatial slot under incomplete binocular evidence. Importantly, the training target is aligned by slot, not by disparity shifted correspondence. The encoder is therefore encouraged to represent what should be present at each spatial location given both views, rather than being directly optimized to produce factorized left right metric descriptors.

\subsection{Stereo aware nuisance model during pretraining}

To make the learned latent useful under realistic stereo conditions, pretraining includes perturbations aligned with the main nuisance factors of correspondence. The two most important are partial visibility and view specific appearance mismatch. We therefore use a stereo aware perturbation recipe that includes occlusion, photometric variation, and noise, with the strongest setting allowing the two views to receive independent appearance perturbations.

These perturbations are not treated as generic regularization. They are paired with the one view masked objective so that the encoder must preserve useful spatial information when one view is incomplete or when the two views differ in appearance. In this sense, the data recipe is part of the representation design: it pushes the encoder toward binocular reasoning under the same failure modes that dominate real stereo.

\subsection{Readout protocols from the same pretrained encoder}

Because the pretrained representation is an integrated binocular latent rather than only a factorized descriptor field, we evaluate it through three readout protocols. These protocols do not alter pretraining. They only expose the same frozen encoder in different ways, each answering a different question about the representation.

\paragraph{Exported descriptor readout.}
To obtain a per view descriptor map without any learned matcher, we evaluate the encoder on the duplicated monocular input $(I,I)$ and average the two phase tokens that correspond to the same patch column:
\[
D(I)_{r,p}
=
\frac{1}{2}\Big(\tilde t_{r,2p}(I,I)+\tilde t_{r,2p+1}(I,I)\Big).
\]
Given descriptor maps $D(L)$ and $D(R)$, stereo is recovered by row wise cosine matching and optional classical semi global matching. This is the strict frozen no linkage descriptor probe used in the main real data results. It asks whether the binocular latent can be exported as a directly usable metric descriptor field.

\paragraph{Shared learned readout.}
A second possibility is that a representation may be binocularly useful even if it is not perfectly factorized into cosine matchable descriptors. To test this fairly, we freeze every encoder and train the same lightweight stereo head for all methods. Since the learned head is identical across models, differences are attributable to the encoder representation rather than to model specific linkage design. This is the shared head benchmark used in the main real data comparison.

\paragraph{Native pair input fused grid readout.}
Finally, we can evaluate the encoder under its intended input mode by feeding the actual stereo pair $(L,R)$ and reading out the fused binocular latent directly. Let
\[
T\in\mathbb{R}^{H_p\times 2W_p\times d}
\]
denote the fused token grid produced by the encoder. A common pair conditioned decoder can then operate directly on this fused grid. This readout preserves the model's early binocular coupling and is especially informative for hard geometry, robustness, and decoder capacity analyses. It asks whether the native fused representation itself makes correspondence easier to decode when the downstream head has direct access to the binocular latent.

\subsection{Interpretation}

The proposed encoder should therefore be understood as a \emph{binocular latent model}. Early binocular fusion injects cross view structure into tokenization itself. Stereo micro cells make that structure local and explicit. The row aware patch phase parameterization imposes an epipolar prior without overcommitting to absolute fused column identity. One view masked token only distillation trains the model to infer slot wise latent scene state from partial binocular evidence, and stereo aware nuisances force that latent to remain useful under occlusion and photometric mismatch.

The three evaluation protocols later used in the experiments then answer complementary questions. The exported descriptor probe asks whether the latent is easy to read out as a direct descriptor field without any learned matcher. The shared head benchmark asks whether the same latent remains strong when every method is granted the same downstream stereo head. The native pair input analyses ask whether the fused binocular representation is especially helpful in the hard regimes where cross view reasoning matters most. This interpretation matches the overall experimental organization of the paper.

\section{Experiments}

We organize the experiments to mirror the paper's central claim. We begin with a controlled benchmark that isolates dual view spatial understanding and identifies which parts of the proposed encoder centric design are causally responsible for strong correspondence behavior. We then move to a strict from scratch real stereo setting on KITTI and evaluate the representation in two complementary ways: a frozen no linkage descriptor probe and a shared head benchmark in which every encoder is paired with the same lightweight stereo matcher. Finally, because the proposed representation is a binocular latent rather than only a factorized descriptor field, we analyze the encoder under its native pair input mode and conclude with a small parameter controlled refinement study.

\subsection{Controlled benchmark for dual view spatial understanding}

We begin with the controlled benchmark because it isolates the representation learning question before any dataset specific complication. The benchmark uses paired views with known horizontal displacement together with optional occlusion and optional photometric perturbation, and evaluation is performed by nearest neighbor row wise token matching on frozen encoder features. No learned matcher is involved. Since the goal of this subsection is causal identification, every table reports only dual view correspondence metrics; single image behavior is not mixed into these controlled ablations.

All controlled ablations use the scaled development model employed for rapid iteration on the controlled data. As a consequence, the absolute parameter counts in Tables~\ref{tab:ablation_fusion_token}--\ref{tab:ablation_data} differ from those of the final real data model. The purpose here is not to reproduce the final KITTI numbers, but to identify which architectural, objective, and data choices are responsible for strong dual view spatial behavior.

\subsubsection{Architecture: fusion, tokenization, and positional parameterization}

We first ablate how binocular structure enters the encoder, following the same order as the method section: fusion and tokenization first, then positional parameterization.

\begin{table*}[t]
\centering
\small
\setlength{\tabcolsep}{4pt}
\caption{Architecture ablations on the controlled dual-view benchmark. Interleave fusion is clearly preferable to left-right concatenation. The best tokenization is the proposed 4$\times$2 stereo micro-cell per view, implemented here as width~4 on the fused image.}
\label{tab:ablation_fusion_token}
\begin{tabular}{lccccccc}
\toprule
Variant & train\_s & params (M) & Ntok & PCK@0 (\%) & PCK@1 (\%) & PCK@2 (\%) & EPE \\
\midrule
Interleave, stride 1, width 4 & 162.3 & 2.233 & 128 & \textbf{75.96} & \textbf{87.29} & \textbf{91.76} & \textbf{0.557} \\
Concat, width 4 & 109.9 & 2.233 & 128 & 71.19 & 84.64 & 89.92 & 0.670 \\
Interleave, width 2 & 289.6 & 2.229 & 256 & 74.96 & 86.59 & 91.03 & 0.591 \\
Interleave, width 8 & 95.3 & 2.242 & 64 & 73.28 & 86.01 & 90.89 & 0.612 \\
Interleave, width 1 & 350.3 & 2.226 & 512 & 60.24 & 77.28 & 85.08 & 0.959 \\
Interleave, stride 2, width 4 & 160.0 & 2.233 & 128 & 74.00 & 86.48 & 91.39 & 0.587 \\
Interleave, stride 4, width 8 & 94.7 & 2.242 & 64 & 74.24 & 86.32 & 91.07 & 0.595 \\
\bottomrule
\end{tabular}
\end{table*}

Table~\ref{tab:ablation_fusion_token} yields three observations. First, pixel interleave is clearly superior to simple left-right concatenation. At matched token count and parameter count, interleave improves PCK@0 by 4.77 points and reduces EPE by 0.113. Second, the best tokenization is the proposed 4$\times$2 stereo micro-cell per view. Very fine tokenization with width~1 degrades performance sharply, while coarser tokenization with width~8 reduces spatial precision. Third, changing the exact interleave block pattern has only a modest effect, which suggests that the dominant factor is early binocular coupling itself rather than the precise ordering of view fragments.

\begin{table}[t]
\centering
\small
\setlength{\tabcolsep}{4pt}
\caption{Positional encoding ablations under the selected interleave 4$\times$2 architecture. The proposed patch-phase 2D encoding yields the best error profile while using fewer parameters than the 1D baseline.}
\label{tab:ablation_pos}
\begin{tabular}{lcccccc}
\toprule
Variant & params (M) & train\_s & PCK@0 (\%) & PCK@1 (\%) & PCK@2 (\%) & EPE \\
\midrule
1D positional embedding & 2.233 & 161.0 & \textbf{76.07} & 87.37 & 91.80 & 0.553 \\
Factorized 2D row+col & 2.149 & 159.6 & 75.24 & 87.03 & 91.40 & 0.578 \\
Full 2D grid & 2.233 & 159.4 & 75.61 & 87.17 & 91.76 & 0.559 \\
De-interleaved center 2D & 2.149 & 159.4 & 72.97 & 84.51 & 89.51 & 0.674 \\
Patch-phase 2D (ours) & \textbf{2.139} & 159.8 & 75.90 & \textbf{88.76} & \textbf{92.95} & \textbf{0.507} \\
\bottomrule
\end{tabular}
\end{table}

Table~\ref{tab:ablation_pos} then examines positional encoding under the selected interleave 4$\times$2 architecture. Generic two-dimensional encodings are competitive, but the strongest design is the proposed patch-phase encoding aligned with the stereo micro-cell structure. This design yields the best EPE together with the best PCK@1 and PCK@2, while using fewer parameters than the one-dimensional baseline. The PCK@0 difference is negligible. We therefore select interleave fusion with 4$\times$2 stereo micro-cells and patch-phase positional encoding for all subsequent experiments.

\subsubsection{Objective design}

We next hold the architecture fixed and ask which supervisory signal is most compatible with dense dual view reasoning.

\begin{table*}[t]
\centering
\small
\setlength{\tabcolsep}{4pt}
\caption{Objective ablations under the selected architecture. Token-level uniform distillation is best on the primary dual-view spatial benchmark. Global CLS supervision weakens dense matching, and masking both views destroys the intended cross-view completion behavior.}
\label{tab:ablation_objective}
\begin{tabular}{lcccccc}
\toprule
Variant & train\_s & params (M) & PCK@0 (\%) & PCK@1 (\%) & PCK@2 (\%) & EPE \\
\midrule
Token only, uniform, EMA, scheduled mask & 188.5 & 2.139 & \textbf{79.46} & \textbf{90.85} & \textbf{93.68} & \textbf{0.453} \\
Token only, weighted, EMA, scheduled mask & 187.9 & 2.139 & 79.11 & 90.18 & 93.15 & 0.479 \\
Token+CLS, uniform, EMA, scheduled mask & 188.7 & 2.139 & 76.72 & 88.87 & 92.48 & 0.528 \\
Token+CLS, weighted, EMA, scheduled mask & 191.6 & 2.139 & 75.51 & 87.83 & 91.84 & 0.565 \\
Token+CLS, weighted, EMA, fixed mask 0.50 & 173.0 & 2.139 & 74.64 & 87.17 & 91.50 & 0.588 \\
Token+CLS, weighted, EMA, mask both views & 192.6 & 2.139 & 71.07 & 88.33 & 92.04 & 0.600 \\
CLS only, EMA, scheduled mask & 188.3 & 2.139 & 63.67 & 80.60 & 86.94 & 0.865 \\
\bottomrule
\end{tabular}
\end{table*}

Table~\ref{tab:ablation_objective} leads to three conclusions. First, token-level supervision is essential. A pure CLS objective performs poorly, which confirms that global invariance alone is insufficient for dense correspondence. Second, the task must preserve one intact source view. Masking both views degrades performance substantially, which directly supports the intended completion principle: the encoder should recover missing content from the complementary viewpoint. Third, a curriculum on mask ratio is beneficial, as the scheduled mask ratio improves over a fixed ratio.

Most importantly, the best objective on this benchmark is token-only uniform distillation. Removing the CLS term improves correspondence substantially, and removing extra weighting on masked tokens improves it further. For the primary dual-view spatial objective, local token alignment is therefore the dominant supervision signal, whereas the global branch appears to weaken local discrimination.

\subsubsection{Training distribution and robustness}

Finally, we study the training distribution. The goal here is to determine which nuisance factors are required for robust dual view spatial understanding rather than only clean matching.

\begin{table*}[t]
\centering
\small
\setlength{\tabcolsep}{4pt}
\caption{Data and robustness ablations. \texttt{HARD\_S1} is the main target benchmark. \texttt{HARD\_S2} increases the displacement range and acts as an out-of-distribution test. \texttt{EASY\_S1} is a clean reference setting.}
\label{tab:ablation_data}
\begin{tabular}{lcccccccc}
\toprule
Variant & train\_s & params (M) & HARD\_S1 PCK@1 (\%) & HARD\_S1 EPE & HARD\_S2 PCK@1 (\%) & HARD\_S2 EPE & EASY\_S1 PCK@1 (\%) & EASY\_S1 EPE \\
\midrule
Easy only & 176.9 & 1.670 & 89.63 & 0.518 & 86.01 & 0.613 & 95.56 & 0.293 \\
Occlusion only & 175.1 & 1.670 & 90.35 & 0.476 & 87.83 & 0.530 & 95.75 & 0.268 \\
Photometric+noise only, shared & 180.9 & 1.670 & 90.43 & 0.486 & 87.41 & 0.554 & 95.75 & 0.286 \\
Occlusion+photometric, shared & 179.3 & 1.670 & 90.65 & 0.462 & 88.29 & 0.510 & 95.83 & 0.261 \\
Full hard, shared perturbation & 181.9 & 1.670 & 90.79 & 0.456 & 88.27 & 0.509 & 95.85 & 0.261 \\
Full hard, independent perturbation & \textbf{182.0} & \textbf{1.670} & \textbf{91.17} & \textbf{0.448} & \textbf{88.30} & 0.513 & \textbf{96.43} & \textbf{0.247} \\
\bottomrule
\end{tabular}
\end{table*}

The data ablations in Table~\ref{tab:ablation_data} show that hard training conditions improve exactly the regime that matters for the method. Training on clean pairs is sufficient for the easy test but generalizes less well to the hard settings. Adding occlusion alone already yields a substantial gain, which confirms that occlusion reasoning is central to cross-view learning. Photometric variation further improves robustness, and the strongest overall recipe is obtained when the two views receive independent appearance perturbations. This configuration gives the best HARD\_S1 metrics and the best EASY\_S1 metrics, while remaining essentially tied with shared perturbation on HARD\_S2. Importantly, the hardest training recipe does not trade away clean performance; it improves the main hard setting while also improving the easy reference setting.

\subsubsection{Final controlled configuration}

Combining the architectural, objective, and data ablations yields the final controlled configuration used to motivate the real data model: interleave fusion with stride~1, $4\times2$ stereo micro cell tokens, patch phase positional encoding, token only uniform distillation, an EMA teacher with centering, mask one view with a scheduled mask ratio, and the full hard independent training recipe. Under this configuration, the model reaches 79.52 PCK@0, 91.17 PCK@1, 93.82 PCK@2, and 0.448 EPE on the hard dual view benchmark. This controlled stage therefore establishes the causal story of the paper before we move to real stereo: early binocular micro cells, row aware positioning, token only cross view supervision, and hard stereo nuisances are the main ingredients behind the final gains.

\subsection{Experimental protocol on real stereo data}

We next move from the controlled benchmark to a strict from scratch low resource real stereo setting. All encoders are pretrained self supervised on the unlabeled stereo pairs from the KITTI object dataset, which provides 7{,}481 synchronized left right pairs without ground truth disparity. The purpose of this setup is to isolate architectural inductive bias and binocular representation quality rather than to inherit large scale external pretraining. Accordingly, all compared models use the same resized input resolution, token stride, optimizer, and training budget. The comparison includes our proposed encoder \ours, together with scaled but paper faithful implementations of CroCo-v1, CroCo-v2, iBOT style, DINOv2 style, and DINOv3 style. Since this is a small data regime relative to the original large scale self supervised setups, the comparison should be interpreted as a controlled study of data efficiency and stereo representation quality rather than a replication of released checkpoints.

We evaluate the learned representation in two complementary ways. The first is a strict frozen no linkage descriptor probe: dense stereo is recovered exclusively from frozen encoder descriptors using row wise cosine matching and optional classical SGM refinement. This protocol measures how easily the latent can be exported as a direct metric descriptor. The second is a shared head benchmark that addresses the fairness concern that some baselines are usually paired with a learned stereo matcher. In this setting, every encoder is frozen after self supervised pretraining on KITTI object, and the same lightweight stereo head is trained for all methods on KITTI Stereo~2012 ground truth. This benchmark tests the quality of the encoder representation under a common learned readout while removing any advantage from model specific downstream design.

For the frozen no linkage descriptor probe, we use three real data evaluations. First, on KITTI object we construct a proxy dense stereo benchmark by using StereoSGBM as a reference disparity generator. This is not a replacement for real ground truth, but it provides a large real image probe of descriptor quality. Second, we evaluate hard negative left to right retrieval on KITTI object. Third, we evaluate real disparity accuracy on the validation split of KITTI Stereo~2012 using ground truth \texttt{disp\_noc}. In addition to ground truth only metrics, we apply a left right consistency check and report the retained fraction of valid tokens, denoted \texttt{LRkeep}, in order to separate consistency from correctness. As a supplementary transfer check, we also repeat both the frozen no linkage descriptor probe and the shared head benchmark on a deterministic held out split of KITTI Stereo~2015 training scenes, using the same low resource pretraining recipe and \texttt{disp\_occ\_0} for ground truth evaluation.

For the shared head benchmark, the stereo head has 0.086M parameters for every model and consists of shared left right token projections, a groupwise correlation volume, and a small 3D regularization network that predicts disparity logits on the token grid. The training loss is identical for all encoders and combines smooth regression on soft disparity with classification on the rounded target disparity. At evaluation time we report raw winner take all prediction, SGM based local refinement, and the effect of left right consistency filtering.

At the chosen resized resolution, only 0.2\% of KITTI Stereo~2012 validation tokens have disparity larger than 60 pixels. As a result, the standard KITTI criterion $\max(3\text{ px}, 0.05\,d_{\mathrm{gt}})$ numerically collapses to the 3 pixel threshold for almost all evaluation points, making D1 effectively identical to Bad3 in this setting. For this reason, we emphasize EPE and the token aware \texttt{Bad@1tok} metric throughout the main text.

\subsection{Main results on real stereo data}

We now evaluate whether the design choices identified on the controlled benchmark transfer to real stereo.

\subsubsection{Frozen no-linkage descriptor probe}

\begin{table}[t]
\centering
\small
\setlength{\tabcolsep}{5pt}
\caption{Proxy dense stereo on KITTI object using StereoSGBM reference disparities. This experiment provides a large real image probe of descriptor quality without requiring stereo ground truth. Higher \texttt{PCK@1tok} and lower EPE are better.}
\label{tab:kitti_proxy_dense}
\begin{tabular}{lcc}
\toprule
Method & PCK@1tok (\%) & EPE (px) \\
\midrule
\fullours & \textbf{84.07} & \textbf{11.82} \\
CroCo-v1 (sincos) & 52.36 & 18.68 \\
CroCo-v2 (RoPE) & 79.09 & 16.53 \\
iBOT-style & 60.15 & 29.49 \\
DINOv2-style & 60.15 & 29.43 \\
DINOv3-style & 71.55 & 20.11 \\
\bottomrule
\end{tabular}
\end{table}

Table~\ref{tab:kitti_proxy_dense} provides the largest real-image probe of descriptor quality. Although StereoSGBM reference disparities are not human ground truth, the ranking is already highly informative. Our method achieves the best result by a clear margin, reaching 84.07\% \texttt{PCK@1tok} and 11.82 px EPE. Relative to the strongest baseline, CroCo-v2, this corresponds to a gain of 4.98 percentage points in \texttt{PCK@1tok} and a 28.5\% reduction in EPE. By contrast, the DINO and iBOT family is substantially weaker, which suggests that generic self-distillation alone does not produce sufficiently sharp stereo descriptors in this low-resource real-data regime.

\begin{table}[t]
\centering
\small
\setlength{\tabcolsep}{4pt}
\caption{Hard-negative retrieval on KITTI object. Retrieval is evaluated on 1{,}500 left-right pairs. Positives are true left-right matches. Negatives are mined from a hard random subset. Top-1 and \texttt{Hard@1} are the most informative metrics. The mean margin is computed as positive similarity minus the strongest mined negative.}
\label{tab:kitti_retrieval}
\begin{tabular}{lccccc}
\toprule
Method & Top-1 (\%) & Top-5 (\%) & Hard@1 (\%) & Hard@5 (\%) & Margin \\
\midrule
\fullours & \textbf{65.2} & \textbf{82.4} & \textbf{76.1} & \textbf{88.4} & \textbf{+0.0007} \\
CroCo-v1 (sincos) & 45.1 & 69.1 & 59.1 & 79.5 & -0.0015 \\
CroCo-v2 (RoPE) & 56.3 & 78.6 & 69.2 & 85.7 & -0.0011 \\
iBOT-style & 12.4 & 32.9 & 24.1 & 51.1 & -0.0009 \\
DINOv2-style & 14.5 & 32.2 & 25.1 & 49.2 & -0.0040 \\
DINOv3-style & 37.9 & 62.1 & 51.7 & 74.0 & -0.0035 \\
\bottomrule
\end{tabular}
\end{table}

Table~\ref{tab:kitti_retrieval} evaluates pair-level dual-view consistency through hard-negative retrieval. Under this protocol, our method again achieves the best result, with 65.2\% Top-1 and 76.1\% \texttt{Hard@1}. Relative to CroCo-v2, this is an improvement of 8.9 percentage points in Top-1 and 6.9 percentage points in \texttt{Hard@1}. Importantly, our method is also the only model with a positive average hard-negative margin. This indicates that the encoder is not only more globally consistent across the two views, but also more selective when the negatives are visually similar.

\begin{table*}[t]
\centering
\small
\setlength{\tabcolsep}{4pt}
\caption{Real ground-truth disparity on KITTI Stereo~2012 validation under the strict frozen no-linkage descriptor probe. \texttt{GT-only} uses all ground-truth-valid tokens. \texttt{GT+LR} uses only tokens that also pass the left-right consistency check. \texttt{SGMLOC} denotes four-direction SGM on the encoder cost volume followed by local soft refinement. Lower EPE and \texttt{Bad@1tok} are better. Higher \texttt{LRkeep} is better.}
\label{tab:kitti_gt_nolinkage}
\begin{tabular}{lcccccc}
\toprule
Method &
GT WTA EPE &
GT SGMLOC EPE &
GT SGMLOC Bad@1tok (\%) &
GT+LR SGMLOC EPE &
GT+LR SGMLOC Bad@1tok (\%) &
LRkeep (\%) \\
\midrule
\fullours & \textbf{9.74} & \textbf{4.62} & \textbf{2.19} & \textbf{4.57} & \textbf{1.91} & 92.5 \\
CroCo-v1 (sincos) & 14.76 & 10.25 & 21.36 & 10.22 & 21.29 & \textbf{96.8} \\
CroCo-v2 (RoPE) & 14.32 & 5.93 & 8.37 & 5.70 & 7.85 & 87.0 \\
iBOT-style & 28.78 & 7.63 & 11.93 & 7.55 & 11.57 & 71.7 \\
DINOv2-style & 34.07 & 8.44 & 15.68 & 8.39 & 15.44 & 67.1 \\
DINOv3-style & 19.53 & 8.01 & 14.22 & 7.98 & 14.03 & 80.9 \\
\bottomrule
\end{tabular}
\end{table*}

The real ground truth evaluation in Table~\ref{tab:kitti_gt_nolinkage} confirms the same ranking as the proxy benchmark. Under this strict no linkage probe, our model attains the best raw WTA EPE and the best refined SGMLOC EPE on ground truth valid tokens. Relative to CroCo-v2, this corresponds to a 32.0\% reduction in raw WTA EPE and a 22.1\% reduction in refined SGMLOC EPE. The gain is even larger on the token aware thresholded error: our method reduces GT only \texttt{Bad@1tok} from 8.37\% to 2.19\%, and after left right consistency filtering further improves to 1.91\% while retaining 92.5\% of valid tokens. This protocol is deliberately a descriptor readout probe rather than the only meaningful interpretation of the latent, but the result is important because it shows that the learned binocular latent is unusually easy to export as a directly usable descriptor field.

\subsubsection{Shared-head benchmark}

\begin{table*}[t]
\centering
\small
\setlength{\tabcolsep}{4pt}
\caption{Shared-head benchmark on KITTI Stereo~2012. Each encoder is frozen after self-supervised pretraining on KITTI object. The same 0.086M-parameter stereo head is trained on KITTI Stereo~2012 training ground truth and evaluated on the validation split. Lower EPE and \texttt{Bad@1tok} are better. Since the head is identical for all encoders, differences are attributable to the encoder representations rather than the matching head.}
\label{tab:kitti_shared_head}
\begin{tabular}{lcccccccc}
\toprule
Method &
Encoder params (M) &
GT WTA EPE &
GT WTA Bad@1tok (\%) &
GT SGMLOC EPE &
GT SGMLOC Bad@1tok (\%) &
GT+LR SGMLOC EPE &
GT+LR SGMLOC Bad@1tok (\%) &
LRkeep (\%) \\
\midrule
\fullours & \textbf{1.339} & 4.423 & 0.78 & \textbf{4.258} & 0.70 & \textbf{4.215} & 0.56 & 99.4 \\
CroCo-v1 (sincos) & 3.337 & 5.095 & 2.99 & 4.759 & 2.65 & 4.422 & 1.64 & 77.6 \\
CroCo-v2 (RoPE) & 3.337 & \textbf{4.300} & \textbf{0.46} & 4.285 & \textbf{0.38} & 4.291 & \textbf{0.38} & \textbf{99.7} \\
iBOT-style & 2.020 & 5.606 & 4.04 & 5.742 & 3.64 & 5.718 & 3.62 & 97.7 \\
DINOv2-style & 2.022 & 5.593 & 4.02 & 6.232 & 3.54 & 6.210 & 3.49 & 99.3 \\
DINOv3-style & 1.928 & 5.684 & 4.63 & 5.806 & 3.65 & 5.803 & 3.65 & 99.4 \\
\bottomrule
\end{tabular}
\end{table*}

The shared head benchmark yields a complementary picture. Once every encoder is paired with the same lightweight learned head, the absolute gaps narrow, which is expected because a common readout can compensate for representations that are not directly metric. CroCo-v2 attains the best WTA EPE and the best thresholded errors, while \ours\ attains the best refined EPE both before and after left right consistency filtering. Concretely, \ours\ reaches 4.258 GT SGMLOC EPE and 4.215 GT+LR SGMLOC EPE, compared with 4.285 and 4.291 for CroCo-v2. Most importantly for the paper's main claim, \ours\ remains essentially at parity with the strongest baseline while using only 1.339M encoder parameters, compared with 3.337M for CroCo-v2. This indicates that the proposed encoder learns a compact binocular latent that stays highly competitive even when a common downstream matcher is introduced. We also performed a supplementary KITTI Stereo~2015 shared head transfer experiment under the same frozen encoder pair with common head protocol. The qualitative pattern remains the same: \ours\ and CroCo-v2 again form the strongest pair, with CroCo-v2 slightly better on raw WTA and thresholded \texttt{Bad@1tok}, while \ours\ attains the best refined EPE and does so with a much smaller encoder. We therefore interpret the KITTI Stereo~2015 shared head result as corroborative evidence for the same encoder centric conclusion rather than as a replacement for the main KITTI Stereo~2012 table.

Taken together, Tables~\ref{tab:kitti_proxy_dense}--\ref{tab:kitti_shared_head} show a consistent pattern. In the strict no linkage descriptor probe, \ours\ is the strongest among all compared baselines on proxy dense stereo, hard negative retrieval, and real KITTI Stereo~2012 disparity. Under the shared head benchmark, \ours\ remains one of the two strongest encoders while being substantially smaller than CroCo-v2. These results establish the main real data claim of the paper on KITTI Stereo~2012. Supplementary KITTI Stereo~2015 transfer experiments show the same qualitative trend in both the frozen no linkage and shared head settings.

\subsection{Native pair input analyses}

The results above already establish the paper's main claim under both strict descriptor readout and a shared learned matcher. A remaining question, however, is whether the advantage of \ours\ persists when the model is evaluated under its native binocular input formulation rather than through the duplicated input export pathway used by the strict descriptor probe. We therefore turn to native pair input analyses.

\subsubsection{Native pair input stress test on hard geometry}

\paragraph{Motivation.}
The principal limitation of the earlier encoder only stereo probes is that they partly rely on a monocular export pathway, whereas the proposed architecture is explicitly designed to process a \emph{binocular} input. To test whether the learned representation remains advantageous under its native input distribution, we introduce a fused grid benchmark that operates directly on actual stereo pairs $(L,R)$. We focus on the large disparity regime because this is the part of stereo matching where dual view spatial understanding matters most. When disparity is small, a same column prior and local appearance continuity already provide a strong cue. When disparity is large, however, the correct match lies farther from the trivial alignment, local ambiguity increases, and any weakness in cross view reasoning becomes more exposed. We therefore define a large disparity subset as the top quartile of token disparities on the KITTI Stereo~2012 validation split and treat this regime as a targeted stress test of binocular spatial reasoning.

\paragraph{Experimental setup and implementation.}
All encoders are first pretrained self supervised on unlabeled KITTI object stereo pairs and then frozen. We attach the same lightweight pair conditioned fused grid decoder to every method, with 0.687M parameters and identical optimization, training split, and supervision. For \ours, the decoder consumes the native fused token grid obtained from the actual pair input $(L,R)$, which preserves the model's intended early binocular coupling. For all baselines, we construct the same fused grid format by interleaving the exported left and right token maps into a common sequence, ensuring that the downstream decoder architecture is matched across methods. The decoder predicts disparity on the patch column grid, and we report \texttt{Bad@1tok} because it is the most informative token aware criterion in this resized setting. This design isolates the question of interest: whether the frozen pair conditioned representation itself makes large disparity geometry easier to decode under a common downstream head.

\begin{table}[t]
\centering
\small
\setlength{\tabcolsep}{5pt}
\caption{Large-disparity native pair-input stress test on KITTI Stereo~2012 validation. The subset is defined as the top quartile of token disparities on the evaluation grid. All methods use the same 0.687M pair-conditioned fused-grid decoder. \ours\ is evaluated in its native pair-input mode on the actual stereo pair $(L,R)$, while the export control uses the duplicated-input descriptor-export pathway based on $(I,I)$. Lower is better.}
\label{tab:native_hard_geometry}
\begin{tabular}{lc}
\toprule
Method & Large-disparity LOCAL Bad@1tok (\%) \\
\midrule
\ours\ (native fused) & \textbf{15.53} \\
\ours\ (export control) & 21.77 \\
CroCo-v1 (sincos) & 16.29 \\
CroCo-v2 (RoPE) & 17.66 \\
iBOT-style & 20.93 \\
DINOv2-style & 21.23 \\
DINOv3-style & 17.81 \\
\bottomrule
\end{tabular}
\end{table}

\paragraph{Result and interpretation.}
Table~\ref{tab:native_hard_geometry} shows that \ours\ attains the lowest large disparity \texttt{Bad@1tok}, at 15.53\%. Relative to the export based control, native pair input evaluation reduces large disparity error by 6.24 percentage points, corresponding to a 28.7\% relative reduction. This is an important result because it shows that the proposed architecture is not merely compatible with the binocular input distribution on which it is trained; it benefits from it substantially in the most geometrically challenging regime. The same native pair input model also improves over the strongest competing baselines on this hard slice, reducing \texttt{Bad@1tok} by 0.76 points relative to CroCo-v1 and by 2.13 points relative to CroCo-v2, while the gains are larger still over the generic self distillation baselines. In other words, once evaluation is restricted to the regime where true cross view reasoning is indispensable, the advantage of the proposed binocular representation becomes most visible.

\paragraph{Why this experiment matters.}
This stress test complements the earlier no linkage and shared head evaluations in an important way. The frozen descriptor probe establishes that \ours\ produces the strongest descriptors under a strict encoder only stereo protocol. The shared head benchmark shows that the representation remains highly competitive when every method is granted the same learned stereo head. The present experiment adds a third piece of evidence: when the model is evaluated under its native pair input formulation, and the benchmark is concentrated on the hard geometry regime rather than easy near column matches, \ours\ yields the strongest performance. We therefore interpret the combined evidence as indicating that the gain of \ours\ is not tied to a single evaluation recipe. Instead, its advantage is most pronounced precisely where a good stereo encoder should matter most, namely in difficult correspondence regions that require stronger binocular spatial understanding.

\subsubsection{Robustness under unilateral one view occlusion}

A second native input question is whether the learned local representation remains readable when one view becomes partially unavailable. This benchmark is especially well aligned with the training principle of the model, since one view masked self distillation explicitly encourages the encoder to preserve useful spatial information when one view is incomplete and the other must supply the missing context.

Each encoder is frozen, and the same weak local 2D head is trained for all methods on the clean KITTI Stereo~2012 training split. The head has only 0.242M parameters and contains no cost volume construction, no 3D regularization, and no model specific linkage design. For the single view baselines, the head receives the paired local feature constructed from the frozen left and right token maps at the same spatial cell. For \ours, the actual stereo pair $(L,R)$ is passed through the frozen binocular encoder, and the fused token sequence is de interleaved into its two phase maps; the same local feature template and the same readout head are then applied. The head is trained only on clean stereo pairs and is never exposed to corrupted inputs during training. At test time, one view is corrupted by token aligned rectangular block occlusion with area ratio from 10\% to 40\%.

\begin{table*}[t]
\centering
\small
\setlength{\tabcolsep}{4pt}
\caption{Native pair-input local disparity readability under unilateral one-view occlusion. The same weak local 2D head (0.242M parameters) is trained on clean KITTI Stereo~2012 training pairs for all methods. At test time, one view is corrupted by token-aligned block occlusion with area ratio from 10\% to 40\%. For \ours, the actual stereo pair $(L,R)$ is passed through the binocular encoder and the fused token sequence is de-interleaved into its two phase maps. Lower is better. \texttt{OccAUC} is the average across the nonzero corruption severities, and \texttt{$\Delta$40} is the degradation from clean to 40\% occlusion.}
\label{tab:native_occlusion}
\begin{tabular}{lcccccccc}
\toprule
Method &
Clean EPE &
Clean Bad@1tok (\%) &
Occ40 EPE &
Occ40 Bad@1tok (\%) &
OccAUC EPE &
OccAUC Bad@1tok (\%) &
$\Delta$40 EPE &
$\Delta$40 Bad@1tok (\%) \\
\midrule
\ours & 5.403 & 4.37 & \textbf{6.886} & \textbf{9.74} & \textbf{6.284} & \textbf{7.50} & \textbf{1.483} & \textbf{5.37} \\
CroCo-v1 (sincos) & 5.360 & 4.70 & 7.081 & 9.88 & 6.366 & 7.89 & 1.721 & 5.18 \\
CroCo-v2 (RoPE) & \textbf{5.028} & \textbf{3.53} & 8.741 & 19.74 & 7.380 & 13.71 & 3.713 & 16.22 \\
iBOT-style & 5.831 & 5.06 & 10.725 & 25.71 & 9.032 & 18.39 & 4.894 & 20.65 \\
DINOv2-style & 5.855 & 5.21 & 8.625 & 17.97 & 7.619 & 13.46 & 2.770 & 12.76 \\
DINOv3-style & 5.562 & 4.58 & 8.928 & 18.69 & 8.218 & 16.25 & 3.365 & 14.11 \\
\bottomrule
\end{tabular}
\end{table*}

Table~\ref{tab:native_occlusion} yields the clearest native pair robustness evidence in the paper. On clean local readout, \ours\ is competitive but not the best model: CroCo-v2 attains lower clean EPE (5.028 vs.\ 5.403 px) and lower clean \texttt{Bad@1tok} (3.53\% vs.\ 4.37\%). This is important because it shows that the benchmark is not biased toward our method and that we do not dominate every operating regime. However, once one view becomes partially missing, the ranking changes sharply. At 40\% one view occlusion, \ours\ reaches 6.886 px EPE and 9.74\% \texttt{Bad@1tok}, whereas CroCo-v2 degrades to 8.741 px and 19.74\%. Averaged across all nonzero corruption levels, \ours\ reduces \texttt{OccAUC} EPE from 7.380 to 6.284 relative to CroCo-v2 and reduces \texttt{OccAUC Bad@1tok} from 13.71\% to 7.50\%. The degradation analysis is even more revealing: from clean input to 40\% occlusion, \ours\ increases by only 1.483 px in EPE and 5.37 points in \texttt{Bad@1tok}, whereas CroCo-v2 increases by 3.713 px and 16.22 points. We interpret this as direct evidence that the proposed architecture learns a more robust binocular local representation, exactly as expected from one view masked self distillation.

\subsubsection{Readout under scalable decoder capacity}

The two native input comparisons above show that the frozen representation remains advantageous under actual binocular input. We finally analyze \ours\ in isolation to ask how much decoder capacity is required to extract geometry from the same fused binocular latent.

We freeze the pretrained encoder and evaluate it on the actual stereo pair $(L,R)$ rather than on the duplicated input export control $(I,I)$. Let
\[
T \in \mathbb{R}^{H_p \times 2W_p \times d}
\]
denote the fused token grid produced by the encoder. The readout head receives this fused grid and predicts disparity on the patch column lattice of width $W_p$. We consider a family of heads with increasing capacity. The weakest head is a linear readout that first collapses each adjacent token pair $(T_{r,2p},T_{r,2p+1})$ into a patch column feature
\[
u_{r,p} = \big[T_{r,2p},\,T_{r,2p+1},\,|T_{r,2p}-T_{r,2p+1}|,\,T_{r,2p}\odot T_{r,2p+1}\big],
\]
and then predicts disparity independently at each spatial location. Stronger heads add 2D convolutional processing either before pair collapse, after pair collapse, or both, thereby allowing the decoder to aggregate local evidence across neighboring fused cells. All heads are trained on the KITTI Stereo~2012 training split with the same loss, optimizer, and resolution. Since the encoder is frozen, all differences are attributable to how much decoder capacity is required to extract geometry from the same native binocular representation.

We report local correspondence ranking metrics (Top-1, Top-3, and MRR) together with the disparity metrics derived from the same logits. The main criterion is the \texttt{LOCAL} prediction, obtained by a local soft readout around the discrete disparity maximum. We also report performance on two more difficult subsets: \texttt{BND}, which corresponds to disparity boundary regions defined by the local disparity gradient, and \texttt{LARGE}, which corresponds to the upper quartile of disparity magnitudes. Coverage is 100\% for all head sizes and is omitted from the table for brevity.

\begin{table*}[t]
\centering
\small
\setlength{\tabcolsep}{4pt}
\caption{Native pair-input \ours\ under increasing decoder capacity. The encoder is frozen and evaluated on the actual stereo pair $(L,R)$. Lower EPE and \texttt{Bad@1tok} are better. Higher Top-1, Top-3, and MRR are better.}
\label{tab:native_scale}
\begin{tabular}{lcccccccccc}
\toprule
Head & Params (M) & Top-1 (\%) & Top-3 (\%) & MRR &
LOCAL EPE & LOCAL Bad@1tok (\%) &
BND EPE & BND Bad@1tok (\%) &
LARGE EPE & LARGE Bad@1tok (\%) \\
\midrule
Linear          & 0.032 & 69.1 & 97.6 & 0.830 & 6.381 & 6.23 & 6.676 & 8.46 & 9.668 & 22.30 \\
Tiny64$\times$1$\times$1   & 0.106 & 78.6 & 97.9 & 0.881 & 5.745 & 5.67 & 6.218 & 8.14 & 8.953 & 21.39 \\
Small96$\times$1$\times$2  & 0.309 & 77.3 & 97.8 & 0.873 & 5.689 & 5.52 & 6.094 & 7.90 & 8.670 & 21.08 \\
Medium128$\times$2$\times$2& 0.687 & 78.2 & 98.9 & 0.880 & 5.591 & 5.04 & 5.972 & 7.22 & \textbf{8.236} & \textbf{19.41} \\
Large192$\times$3$\times$3 & 2.186 & 80.0 & 99.0 & 0.891 & 5.604 & 5.44 & 5.990 & 7.75 & 8.598 & 20.93 \\
XLarge256$\times$4$\times$4& 5.045 & \textbf{80.4} & 98.9 & \textbf{0.893} & \textbf{5.428} & \textbf{5.03} & \textbf{5.813} & \textbf{7.22} & 8.472 & 19.71 \\
\bottomrule
\end{tabular}
\end{table*}

Table~\ref{tab:native_scale} shows a clear scaling trend. As decoder capacity increases, both local ranking quality and disparity accuracy improve substantially. From the weakest linear readout to the best overall configuration, Top-1 rises from 69.1\% to 80.4\%, while MRR improves from 0.830 to 0.893. In terms of disparity quality, \texttt{LOCAL} EPE decreases from 6.381 to 5.428 pixels, and \texttt{Bad@1tok} improves from 6.23\% to 5.03\%. The same pattern holds on difficult spatial subsets, including disparity boundaries and large disparity regions.

This scaling behavior is consistent with the architecture. Because the encoder uses interleaved binocular micro cells and shared self attention, disparity is not expected to emerge as a trivially readable scalar attached to each fused token pair. Instead, the representation is likely to encode stereo geometry in a relational and distributed form. The capacity sweep therefore supports the intended interpretation of the model: early binocular coupling creates a useful pair conditioned spatial latent, but stronger local decoding is increasingly able to recover that geometry more sharply than a minimal linear probe. At the same time, the gains are not fully monotonic. Most of the improvement occurs between the linear and medium scale heads, while the medium scale head remains strongest on the large disparity subset. This suggests that the main limitation of weak heads is insufficient local integration, whereas very large heads may mildly oversmooth rare high disparity cases.

\subsection{Parameter controlled refinements around the real data recipe}

In addition to the main architecture and controlled ablations above, we performed a small parameter controlled study around the real data training recipe in order to test whether lightweight backbone or readout refinements can improve the accuracy invariance trade off. All variants keep the same binocular tokenization, the same self supervised training recipe, the same data, and the same resized input resolution. To keep the comparison fair, the total parameter count is held approximately constant by adjusting the feed forward expansion ratio whenever an additional component is introduced. The resulting models all remain within a narrow budget of 1.77M--1.80M parameters. We test three modifications: adding register tokens, replacing the standard feed forward block with SwiGLU, and introducing separate token global heads with a lightweight global self distillation branch.

\begin{table*}[t]
\centering
\small
\setlength{\tabcolsep}{4pt}
\caption{Parameter-controlled refinements around the real-data recipe. All variants keep approximately the same total parameter count by compensating the feed-forward ratio. Lower stereo error and higher retrieval scores are better.}
\label{tab:extra_ablation}
\begin{tabular}{lcccccccccc}
\toprule
Variant &
Total params (M) &
GT WTA EPE &
GT WTA Bad@1tok (\%) &
GT SGMLOC EPE &
GT SGMLOC Bad@1tok (\%) &
GT+LR SGMLOC EPE &
GT+LR SGMLOC Bad@1tok (\%) &
Top-1 (\%) &
Hard@1 (\%) &
Margin \\
\midrule
Base model &
1.771 &
5.184 &
14.66 &
2.202 &
1.73 &
2.192 &
1.64 &
64.3 &
76.3 &
0.0026 \\

+ register tokens &
1.772 &
5.134 &
14.44 &
2.266 &
1.88 &
2.256 &
1.78 &
58.9 &
72.8 &
0.0015 \\

+ SwiGLU FFN &
1.779 &
\textbf{5.003} &
\textbf{13.83} &
\textbf{2.144} &
\textbf{1.43} &
\textbf{2.131} &
\textbf{1.33} &
64.2 &
76.9 &
0.0028 \\

+ dual token/global heads &
1.797 &
4.450 &
12.38 &
2.308 &
2.01 &
2.300 &
1.96 &
\textbf{70.8} &
\textbf{82.8} &
\textbf{0.0034} \\
\bottomrule
\end{tabular}
\end{table*}

Table~\ref{tab:extra_ablation} yields three conclusions. First, adding register tokens does not help in the present setting. Although the parameter increase is negligible, refined stereo accuracy becomes slightly worse and retrieval also degrades. This suggests that the current binocular tokenization already provides a sufficiently structured latent, and the extra global memory slots do not contribute useful information for either dense correspondence or dual view retrieval.

Second, replacing the standard feed forward block with SwiGLU is the only modification that improves the model consistently on the primary stereo metrics. Relative to the base model, the SwiGLU variant reduces GT SGMLOC EPE from 2.202 to 2.144 and improves GT SGMLOC \texttt{Bad@1tok} from 1.73\% to 1.43\%. The same trend holds after left right consistency filtering, while retrieval remains essentially unchanged. We therefore regard SwiGLU as the most useful lightweight refinement of the base encoder.

Third, introducing separate token global heads together with a lightweight global branch changes the trade off rather than producing a uniform gain. This variant substantially improves retrieval, increasing Top-1 from 64.3\% to 70.8\% and \texttt{Hard@1} from 76.3\% to 82.8\%, while also improving the average retrieval margin. However, its refined stereo performance is slightly worse than the base model. We therefore interpret the dual head design as a reasonable alternative when downstream tasks prioritize pair level consistency or retrieval, but not as the preferred default for dense stereo matching.

Overall, this final study suggests that the main model is already close to a favorable local optimum. Small global memory modifications do not help, stronger global supervision improves retrieval at the expense of refined stereo accuracy, and the only robust improvement comes from upgrading the feed forward block to SwiGLU.

\subsection{Photometric methods and adjacent methods}

We do not include classical photometric reprojection based self supervised stereo baselines as primary comparators because they address a different problem from the one studied in this paper. Such methods are designed for end to end disparity learning under image reconstruction, warping, smoothness, and occlusion handling objectives, whereas our goal is to learn a compact, reusable encoder whose binocular structure is directly reflected in the latent representation and remains usable under frozen encoder evaluation. In a photometric framework, strong performance may arise from the interaction between encoder, disparity decoder, reprojection loss, and regularization, which makes it difficult to isolate encoder quality. Retaining the photometric decoder would therefore violate our frozen encoder and shared head comparison protocol, while removing it would no longer evaluate the method in the regime for which it was designed. We therefore focus our primary comparisons on representation learning baselines rather than end to end photometric stereo pipelines.

Similarly, although DUSt3R and MASt3R are important adjacent methods, they are not direct SSL encoder baselines because they are trained to regress or refine pairwise 3D or matching structure directly. We therefore treat them as out of category references rather than primary baselines.

\section{Mechanistic Analysis of Emergent Binocular Geometry}
\label{sec:mechanistic_geometry}

The preceding sections evaluate the original fused DePos model mainly through downstream performance. To better understand what the architecture actually represents, we now analyze the \emph{internal geometry} of its fused token space without any retraining. All experiments in this section use the frozen encoder and operate directly on token similarities. We study two questions. First, does epipolar structure emerge across depth inside the fused representation? Second, is this structure causally dependent on the complementary view and its horizontal ordering, rather than on a trivial same row bias or absolute position shortcut?

We analyze two datasets. The first is a controlled synthetic stereo benchmark constructed from KITTI object images by applying known horizontal shifts to reflect padded crops, optionally with occlusion and photometric perturbation. The second is KITTI Stereo~2012 validation with real disparity ground truth. For an even phase query token, we define \texttt{RowConc} as the total probability mass assigned to odd phase candidates on the same row, \texttt{GT@0} as the probability mass at the exact ground truth correspondence, \texttt{GT@1} as the mass within a $\pm1$ token window around the ground truth, \texttt{MRR} as the reciprocal rank of the true match, and \texttt{Entropy} as the entropy of the same row correspondence distribution. At the resolution used here, the phase specific grid is $12\times 40$, so the chance level for \texttt{RowConc} is $1/12 \approx 0.083$, while the chance level for \texttt{GT@1} is approximately $3/40 = 0.075$.

\subsection{Layerwise emergence of epipolar structure}

\begin{figure*}[t] 
\centering 
\includegraphics[width=\textwidth]{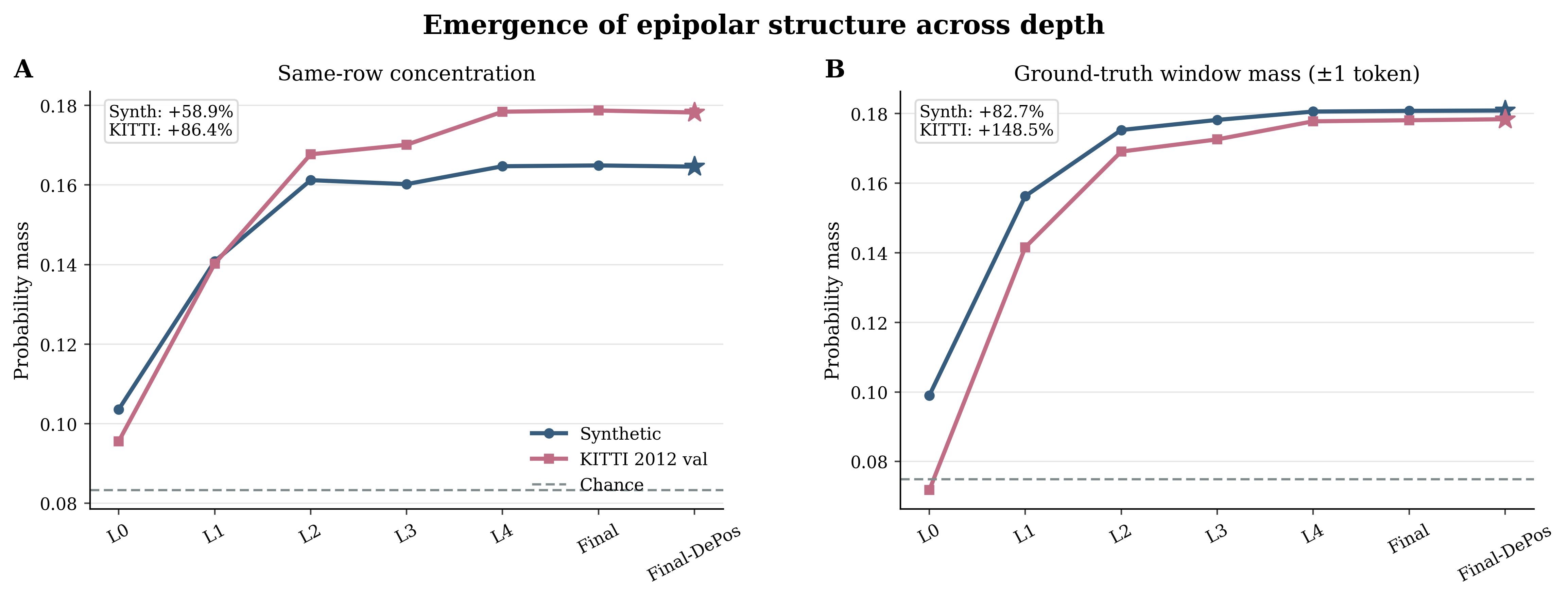} \caption{\textbf{Emergence of epipolar structure across depth in the fused DePos representation.} Same-row concentration and ground-truth window mass increase steadily from the embedded input state to the final DePos readout on both controlled synthetic stereo and KITTI Stereo~2012 validation. Dashed lines indicate chance levels. The final DePos readout is marked explicitly to show that applying DePos preserves the already-emergent geometry rather than altering it substantially.} 
\label{fig:emergent_epipolar} 
\end{figure*}

Figure~\ref{fig:emergent_epipolar} visualizes the main trend, while Table~\ref{tab:layerwise_epipolar} reports the exact values. The fused DePos representation becomes progressively more epipolar structured across depth. On the controlled synthetic benchmark, \texttt{RowConc} rises from 0.1036 at the embedded input state to 0.1646 at the final DePos readout, while \texttt{GT@1} rises from 0.0990 to 0.1809 and \texttt{MRR} rises from 0.2200 to 0.2852. At the same time, entropy drops from 3.4568 to 2.9446. The same trend appears on real KITTI Stereo~2012 validation: \texttt{RowConc} increases from 0.0956 to 0.1782, \texttt{GT@1} from 0.0718 to 0.1784, and \texttt{MRR} from 0.3875 to 0.4283, while entropy decreases from 3.5953 to 2.9658. Relative to the embedded input state, this corresponds to an 86.4\% increase in same row concentration and a 148.5\% increase in ground truth window mass on KITTI. These results indicate that the architecture does not merely preserve low level similarity across layers; instead, it organizes the fused token geometry increasingly around stereo compatible structure.

A second observation is that the final raw representation and the final DePos readout are numerically almost identical on all geometry metrics. For example, on KITTI Stereo~2012, \texttt{RowConc} changes only from 0.1787 to 0.1782 and \texttt{GT@1} changes only from 0.1781 to 0.1784 after applying DePos. This suggests that DePos does not create the epipolar structure by itself; rather, the structure already exists in the fused representation, and DePos preserves it while removing explicit positional content from the exported features. Finally, exact argmax accuracy does not improve monotonically on KITTI despite the gains in \texttt{GT@1}, \texttt{MRR}, and entropy. We interpret this as evidence that the model encodes disparity in a \emph{distributed} local geometry rather than as a purely nearest neighbor descriptor space.

\begin{table*}[t]
\centering
\small
\setlength{\tabcolsep}{4pt}
\caption{Layerwise emergence of epipolar structure in the original fused DePos representation. Metrics are computed from frozen token similarities between even-phase query tokens and odd-phase candidate tokens. Higher \texttt{RowConc}, \texttt{GT@1}, and \texttt{MRR} are better. Lower \texttt{Entropy} is better. For brevity, the table reports the final DePos readout rather than both final raw and final DePos; the two are numerically almost identical, as discussed in the text.}
\label{tab:layerwise_epipolar}
\begin{tabular}{lcccccccc}
\toprule
& \multicolumn{4}{c}{Synthetic stereo} & \multicolumn{4}{c}{KITTI Stereo 2012 val} \\
\cmidrule(lr){2-5}\cmidrule(lr){6-9}
Layer & \texttt{RowConc} & \texttt{GT@1} & \texttt{MRR} & \texttt{Entropy} & \texttt{RowConc} & \texttt{GT@1} & \texttt{MRR} & \texttt{Entropy} \\
\midrule
\texttt{layer0\_raw}   & 0.1036 & 0.0990 & 0.2200 & 3.4568 & 0.0956 & 0.0718 & 0.3875 & 3.5953 \\
\texttt{layer1\_raw}   & 0.1408 & 0.1563 & 0.2642 & 2.9820 & 0.1402 & 0.1416 & 0.4406 & 3.1353 \\
\texttt{layer2\_raw}   & 0.1612 & 0.1753 & 0.2760 & 2.8855 & 0.1677 & 0.1691 & 0.4317 & 2.9826 \\
\texttt{layer3\_raw}   & 0.1602 & 0.1782 & 0.2823 & 2.9319 & 0.1701 & 0.1726 & 0.4361 & 2.9892 \\
\texttt{layer4\_raw}   & 0.1647 & 0.1806 & 0.2852 & 2.9476 & 0.1784 & 0.1778 & 0.4285 & 2.9695 \\
\texttt{final\_depos}  & 0.1646 & 0.1809 & 0.2852 & 2.9446 & 0.1782 & 0.1784 & 0.4283 & 2.9658 \\
\bottomrule
\end{tabular}
\end{table*}

\subsection{Counterfactual interventions reveal dependence on complementary view ordering}

We next test whether the learned geometry is genuinely caused by the complementary view and its horizontal organization. Table~\ref{tab:counterfactual_geometry} reports the final layer geometry under four conditions. The \emph{original} condition uses the normal stereo pair. \emph{Replace-right} substitutes the right image with an unrelated right image from another sample. \emph{Row-shuffle-right} permutes right image patch columns independently within each row, preserving row marginals and patch appearance while destroying true horizontal ordering. \emph{Duplicate-left} replaces the right image with a copy of the left image and uses a zero disparity target.

\begin{figure*}[t]
    \centering
    \includegraphics[width=\textwidth]{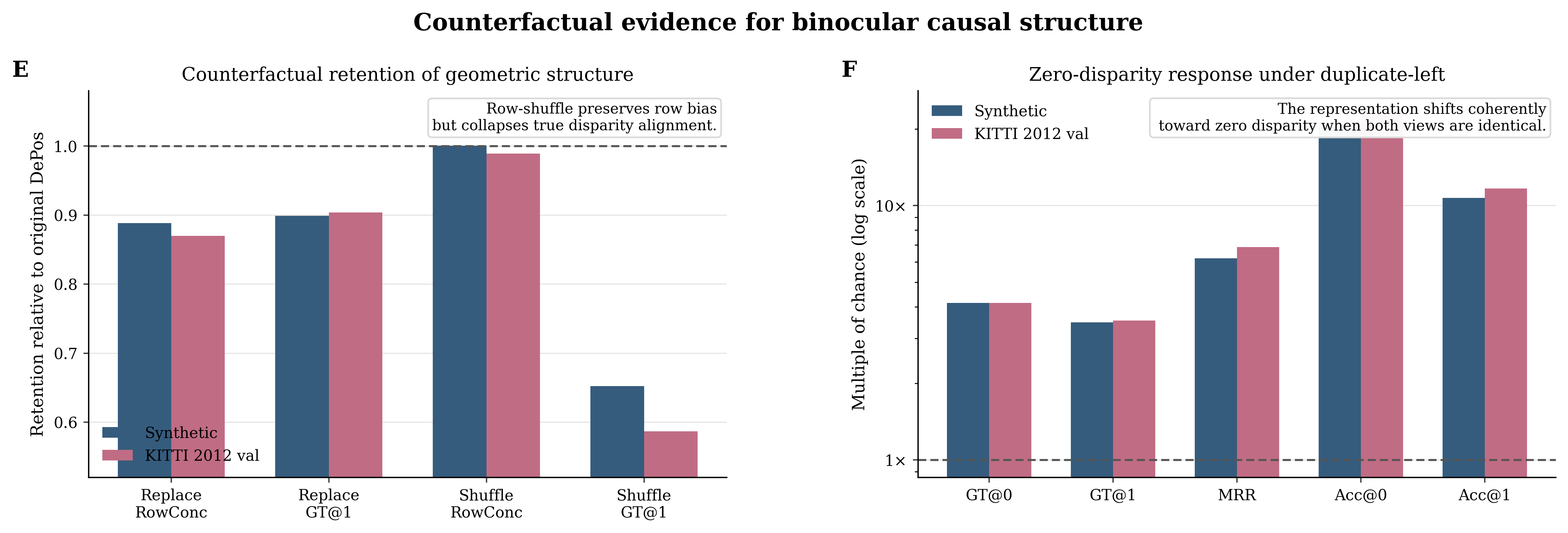}
    \caption{\textbf{Counterfactual evidence for binocular causal structure.}
    \emph{Left:} retention relative to the original DePos readout under \emph{replace-right} and \emph{row-shuffle-right}. Row shuffling preserves same-row concentration but sharply reduces ground-truth-window retention, showing that the learned representation depends on the ordered horizontal structure of the complementary view rather than on a generic row prior. \emph{Right:} under the \emph{duplicate-left} counterfactual, zero-disparity metrics rise to several times chance, indicating that the representation shifts coherently toward zero disparity when the two views are made identical.}
    \label{fig:counterfactual_geometry}
\end{figure*}

Figure~\ref{fig:counterfactual_geometry} summarizes the intervention effects, while Table~\ref{tab:counterfactual_geometry} reports the exact values. The strongest intervention is \emph{row-shuffle-right}. On KITTI Stereo~2012 validation, \texttt{RowConc} remains almost unchanged, moving only from 0.1782 to 0.1762, but \texttt{GT@1} drops from 0.1784 to 0.1047 and \texttt{MRR} drops from 0.4283 to 0.2607, while entropy rises from 2.9658 to 3.2632. The same pattern appears on the synthetic benchmark, where \texttt{GT@1} falls from 0.1783 to 0.1163 and \texttt{MRR} falls from 0.2835 to 0.2339. This is an important mechanistic result: the model retains a coarse same row prior under row shuffling, but the ordered within row structure that defines true stereo correspondence collapses. In other words, the learned representation is not merely ``epipolar row biased''; it is sensitive to the correct \emph{horizontal ordering} of the complementary view.

The \emph{replace-right} intervention is weaker. It reduces \texttt{RowConc} and \texttt{GT@1} and increases entropy, but the degradation is much smaller than under row shuffling. We interpret this to mean that replacing the right image preserves some generic driving scene statistics and row structure, whereas row shuffling specifically destroys the horizontal organization needed for disparity. Finally, the \emph{duplicate-left} intervention produces the expected counterfactual behavior. When the two views are made identical, the representation shifts sharply toward zero disparity. On KITTI Stereo~2012, zero target \texttt{GT@1} increases to 0.2650 and \texttt{MRR} increases to 0.7340, with \texttt{Acc@0}=0.6028 and \texttt{Acc@1}=0.8759. The same pattern holds on the synthetic benchmark. This shows that the model does not encode a fixed generic correspondence pattern; rather, it responds systematically to the binocular relation itself.

Taken together, Tables~\ref{tab:layerwise_epipolar} and~\ref{tab:counterfactual_geometry} together with Figures~\ref{fig:emergent_epipolar} and~\ref{fig:counterfactual_geometry} support the following interpretation. The original fused DePos architecture develops an increasingly epipolar structured token geometry across depth. That geometry is not a trivial artifact of same row concentration alone, because it depends strongly on the ordered horizontal structure of the complementary view. At the same time, applying DePos at readout leaves the geometry essentially unchanged, indicating that DePos preserves the learned stereo structure rather than destroying it.

\begin{table*}[t]
\centering
\small
\setlength{\tabcolsep}{4pt}
\caption{Final-layer counterfactual analysis of the original fused DePos representation. The \emph{original} condition uses the true stereo pair and ground-truth disparity target. \emph{Duplicate-left} uses a zero-disparity target because both inputs are identical. Higher \texttt{RowConc}, \texttt{Tgt@1}, and \texttt{MRR} are better. Lower \texttt{Entropy} is better.}
\label{tab:counterfactual_geometry}
\begin{tabular}{lcccccccc}
\toprule
& \multicolumn{4}{c}{Synthetic stereo} & \multicolumn{4}{c}{KITTI Stereo 2012 val} \\
\cmidrule(lr){2-5}\cmidrule(lr){6-9}
Condition & \texttt{RowConc} & \texttt{Tgt@1} & \texttt{MRR} & \texttt{Entropy} & \texttt{RowConc} & \texttt{Tgt@1} & \texttt{MRR} & \texttt{Entropy} \\
\midrule
Original (raw)         & 0.1726 & 0.1782 & 0.2835 & 2.9708 & 0.1787 & 0.1781 & 0.4284 & 2.9672 \\
Original (DePos)       & 0.1722 & 0.1783 & 0.2835 & 2.9695 & 0.1782 & 0.1784 & 0.4283 & 2.9658 \\
Replace-right          & 0.1530 & 0.1603 & 0.2754 & 3.0595 & 0.1550 & 0.1612 & 0.4281 & 3.0712 \\
Row-shuffle-right      & 0.1723 & 0.1163 & 0.2339 & 3.2364 & 0.1762 & 0.1047 & 0.2607 & 3.2632 \\
Duplicate-left (zero)  & 0.1676 & 0.2603 & 0.6638 & 2.8850 & 0.1745 & 0.2650 & 0.7340 & 2.9336 \\
\bottomrule
\end{tabular}
\end{table*}

\appendix

\section{Supplementary transfer check on KITTI Stereo~2015}

We additionally repeated the frozen no-linkage evaluation on a deterministic held-out split of KITTI Stereo~2015 training scenes using the same low-resource pretraining recipe as in the main KITTI Stereo~2012 experiments. The qualitative trend remains similar: \ours\ stays strongest on held-out hard-negative retrieval and attains the best refined no-linkage disparity after classical regularization.

\begin{table}[t]
\centering
\small
\setlength{\tabcolsep}{4pt}
\caption{Supplementary transfer check on KITTI Stereo~2015 using a deterministic held-out split of the public training scenes. All encoders use the same low-resource pretraining recipe as in the main KITTI Stereo~2012 study.}
\label{tab:kitti2015_transfer}
\begin{tabular}{lcccc}
\toprule
Method & Top-1 (\%) & Top-5 (\%) & GT WTA EPE & GT SGM EPE \\
\midrule
\fullours & \textbf{70.0} & \textbf{100.0} & 12.728 & \textbf{6.728} \\
CroCo-v1 (sincos) & 35.0 & 70.0 & \textbf{9.076} & 10.849 \\
CroCo-v2 (RoPE) & 68.3 & 98.3 & 15.400 & 7.605 \\
iBOT-style & 18.3 & 41.7 & 28.327 & 8.151 \\
DINOv2-style & 16.7 & 55.0 & 23.254 & 9.985 \\
DINOv3-style & 55.0 & 90.0 & 15.059 & 8.437 \\
\bottomrule
\end{tabular}
\end{table}

\section{Shared head transfer check on KITTI Stereo~2015}

We additionally repeated the shared-head benchmark on a deterministic held-out split of KITTI Stereo~2015 training scenes using the same low-resource pretraining recipe as in the main KITTI Stereo~2012 experiments. The qualitative pattern remains similar: \ours\ and CroCo-v2 again form the strongest pair, with CroCo-v2 slightly better on raw WTA and thresholded error, while \ours\ attains the best refined shared-head EPE with a substantially smaller encoder.

\begin{table*}[t]
\centering
\small
\setlength{\tabcolsep}{4pt}
\caption{Supplementary shared-head transfer check on KITTI Stereo~2015 using a deterministic held-out split of the public training scenes. Each encoder is frozen after the same low-resource pretraining recipe as in the main KITTI Stereo~2012 study, and the same lightweight stereo head is trained for all methods. Lower EPE and \texttt{Bad@1tok} are better.}
\label{tab:kitti2015_transfer_sharedhead}
\begin{tabular}{lcccccccc}
\toprule
Method &
Encoder params (M) &
GT WTA EPE &
GT WTA Bad@1tok (\%) &
GT SGMLOC EPE &
GT SGMLOC Bad@1tok (\%) &
GT+LR SGMLOC EPE &
GT+LR SGMLOC Bad@1tok (\%) &
LRkeep (\%) \\
\midrule
\fullours & \textbf{1.339} & 4.742 & 0.95 & \textbf{3.888} & 1.29 & \textbf{3.831} & 1.06 & 99.7 \\
CroCo-v1 (sincos) & 3.337 & 5.022 & 1.97 & 4.313 & 2.05 & 4.209 & 1.63 & 98.4 \\
CroCo-v2 (RoPE) & 3.337 & \textbf{4.637} & \textbf{0.71} & 4.224 & \textbf{0.92} & 4.190 & \textbf{0.77} & \textbf{99.7} \\
iBOT-style & 2.020 & 5.316 & 2.58 & 4.599 & 2.58 & 4.598 & 2.57 & 98.9 \\
DINOv2-style & 2.022 & 5.719 & 4.21 & 5.377 & 4.13 & 5.377 & 4.13 & 100.0 \\
DINOv3-style & 1.928 & 5.294 & 2.47 & 6.401 & 3.63 & 6.401 & 3.63 & 100.0 \\
\bottomrule
\end{tabular}
\end{table*}


\begin{thebibliography}{99}

\bibitem{He2022MAE}
He, K., Chen, X., Xie, S., Li, Y., Dollar, P., and Girshick, R., "Masked Autoencoders Are Scalable Vision Learners," in \emph{Proceedings of the IEEE/CVF Conference on Computer Vision and Pattern Recognition (CVPR)} (2022).

\bibitem{Zhou2021iBOT}
Zhou, J., Wei, C., Wang, H., Shen, W., Xie, C., Yuille, A., and Kong, T., "iBOT: Image BERT Pre-Training with Online Tokenizer," in \emph{International Conference on Learning Representations (ICLR)} (2022).

\bibitem{Oquab2023DINOv2}
Oquab, M., Darcet, T., Moutakanni, T., et al., "DINOv2: Learning Robust Visual Features without Supervision," \emph{Transactions on Machine Learning Research} (2024). arXiv:2304.07193.

\bibitem{Simeoni2025DINOv3}
Simeoni, O., Vo, H. V., Seitzer, M., Baldassarre, F., et al., "DINOv3," arXiv:2508.10104 (2025).

\bibitem{Weinzaepfel2022CroCo}
Weinzaepfel, P., Leroy, V., Lucas, T., et al., "CroCo: Self-Supervised Pre-training for 3D Vision Tasks by Cross-View Completion," in \emph{Advances in Neural Information Processing Systems 35 (NeurIPS)} (2022).

\bibitem{Weinzaepfel2023CroCov2}
Weinzaepfel, P., Lucas, T., Leroy, V., et al., "CroCo v2: Improved Cross-view Completion Pre-training for Stereo Matching and Optical Flow," in \emph{Proceedings of the IEEE/CVF International Conference on Computer Vision (ICCV)} (2023).

\bibitem{Liu2020Flow2Stereo}
Liu, P., King, I., Lyu, M., and Xu, J., "Flow2Stereo: Effective Self-Supervised Learning of Optical Flow and Stereo Matching," in \emph{Proceedings of the IEEE/CVF Conference on Computer Vision and Pattern Recognition (CVPR)} (2020).

\bibitem{Huang2022HNet}
Huang, B., Zheng, J.-Q., Giannarou, S., and Elson, D. S., "H-Net: Unsupervised Attention-based Stereo Depth Estimation Leveraging Epipolar Geometry," in \emph{Proceedings of the IEEE/CVF Conference on Computer Vision and Pattern Recognition Workshops (CVPRW)} (2022).

\bibitem{Yuan2021Multiscopic}
Yuan, W., Zhang, Y., Wu, B., et al., "Stereo Matching by Self-Supervision of Multiscopic Vision," in \emph{IEEE/RSJ International Conference on Intelligent Robots and Systems (IROS)} (2021).

\bibitem{Aleotti2020ReversingCycle}
Aleotti, F., Tosi, F., Zhang, L., Poggi, M., and Mattoccia, S., "Reversing the Cycle: Self-Supervised Deep Stereo Through Enhanced Monocular Distillation," in \emph{European Conference on Computer Vision (ECCV)} (2020).

\bibitem{Chen2021ReciprocalStereoMono}
Chen, Z., Ye, X., Yang, W., et al., "Revealing the Reciprocal Relations Between Self-Supervised Stereo and Monocular Depth Estimation," in \emph{Proceedings of the IEEE/CVF International Conference on Computer Vision (ICCV)} (2021).

\bibitem{Zhou2023TiODepth}
Zhou, Z. and Dong, Q., "Two-in-One Depth: Bridging the Gap Between Monocular and Binocular Self-Supervised Depth Estimation," in \emph{Proceedings of the IEEE/CVF International Conference on Computer Vision (ICCV)} (2023).

\bibitem{Wang2024DUSt3R}
Wang, S., Leroy, V., Cabon, Y., Chidlovskii, B., and Revaud, J., "DUSt3R: Geometric 3D Vision Made Easy," in \emph{Proceedings of the IEEE/CVF Conference on Computer Vision and Pattern Recognition (CVPR)} (2024).

\bibitem{Leroy2024MASt3R}
Leroy, V., Cabon, Y., and Revaud, J., "Grounding Image Matching in 3D with MASt3R," in \emph{European Conference on Computer Vision (ECCV)} (2024).

\bibitem{Tschernezki2022N3F}
Tschernezki, V., Laina, I., Larlus, D., and Vedaldi, A., "Neural Feature Fusion Fields: 3D Distillation of Self-Supervised 2D Image Representations," in \emph{Proceedings of the International Conference on 3D Vision (3DV)} (2022).

\bibitem{Ding2022KDMVS}
Ding, Y., Zhu, Q., Liu, X., Yuan, W., Zhang, H., and Zhang, C., "KD-MVS: Knowledge Distillation Based Self-supervised Learning for Multi-view Stereo," in \emph{European Conference on Computer Vision (ECCV)} (2022).

\bibitem{Shi2023ThreeDDistillation}
Shi, X., Dikov, G., Reitmayr, G., Kim, T.-K., and Ghafoorian, M., "3D Distillation: Improving Self-Supervised Monocular Depth Estimation on Reflective Surfaces," in \emph{Proceedings of the IEEE/CVF International Conference on Computer Vision (ICCV)} (2023).

\bibitem{Tosi2023NeRFSupervisedStereo}
Tosi, F., Tonioni, A., De Gregorio, D., and Poggi, M., "NeRF-Supervised Deep Stereo," in \emph{Proceedings of the IEEE/CVF Conference on Computer Vision and Pattern Recognition (CVPR)} (2023).

\bibitem{Hirschmuller2008SGM}
Hirschmueller, H., "Stereo Processing by Semiglobal Matching and Mutual Information," \emph{IEEE Transactions on Pattern Analysis and Machine Intelligence} \textbf{30}(2), 328--341 (2008).

\bibitem{Wen2025FoundationStereo}
Wen, B., Trepte, M., Aribido, J., Kautz, J., Gallo, O., and Birchfield, S., "FoundationStereo: Zero-Shot Stereo Matching," in \emph{Proceedings of the IEEE/CVF Conference on Computer Vision and Pattern Recognition (CVPR)} (2025).

\bibitem{Geiger2012KITTI}
Geiger, A., Lenz, P., and Urtasun, R., "Are We Ready for Autonomous Driving? The KITTI Vision Benchmark Suite," in \emph{Proceedings of the IEEE Conference on Computer Vision and Pattern Recognition (CVPR)} (2012).

\end{thebibliography}
\end{document}